\documentclass[lettersize,journal]{IEEEtran}
\usepackage{amsmath,amsfonts}
\usepackage{algorithmic}
\usepackage{algorithm}
\usepackage{array}
\usepackage[caption=false,font=normalsize,labelfont=sf,textfont=sf]{subfig}
\usepackage{textcomp}
\usepackage{stfloats}
\usepackage{url}
\usepackage{verbatim}
\usepackage{graphicx}
\usepackage{cite}
\usepackage{microtype}
\usepackage{booktabs} 

\usepackage{amssymb}
\usepackage{mathtools}
\usepackage{amsthm}
\usepackage{comment}

\usepackage{bm}             
\usepackage{bbm}            
\usepackage{multirow}
\usepackage{enumitem}

\begin{document}

\title{Concrete Dense Network for Long-Sequence \\Time Series Clustering}

\author{
\IEEEauthorblockN{Redemptor Jr Laceda Taloma\IEEEauthorrefmark{1},
Patrizio Pisani\IEEEauthorrefmark{2},
Danilo Comminiello\IEEEauthorrefmark{1}}

\IEEEauthorblockA{\IEEEauthorrefmark{1}DIET, Sapienza University of Rome, ITALY Email: \{redemptorjr.taloma, danilo.comminiello\}@uniroma1.it}
\IEEEauthorblockA{\IEEEauthorrefmark{2}Unidata S.p.A., ITALY Email: patrizio.pisani@unidata.it}}



\maketitle

\begin{abstract}
Time series clustering is fundamental in data analysis for discovering temporal patterns. Despite recent advancements, learning cluster-friendly representations is still challenging, particularly with long and complex time series. Deep temporal clustering methods have been trying to integrate the canonical \textit{k}-means into end-to-end training of neural networks but fall back on surrogate losses due to the non-differentiability of the hard cluster assignment, yielding sub-optimal solutions. In addition, the autoregressive strategy used in the state-of-the-art RNNs is subject to error accumulation and slow training, while recent research findings have revealed that Transformers are less effective due to time points lacking semantic meaning, to the permutation invariance of attention that discards the chronological order and high computation cost.
In light of these observations, we present LoSTer which is a novel dense autoencoder architecture for the long-sequence time series clustering problem (LSTC) capable of optimizing the \textit{k}-means objective via the Gumbel-softmax reparameterization trick and designed specifically for accurate and fast clustering of long time series. Extensive experiments on numerous benchmark datasets and two real-world applications prove the effectiveness of LoSTer over state-of-the-art RNNs and Transformer-based deep clustering methods.
\end{abstract}


\section{Introduction}
\begin{figure*}[ht]
  \centering
  \includegraphics[width=0.97\textwidth]{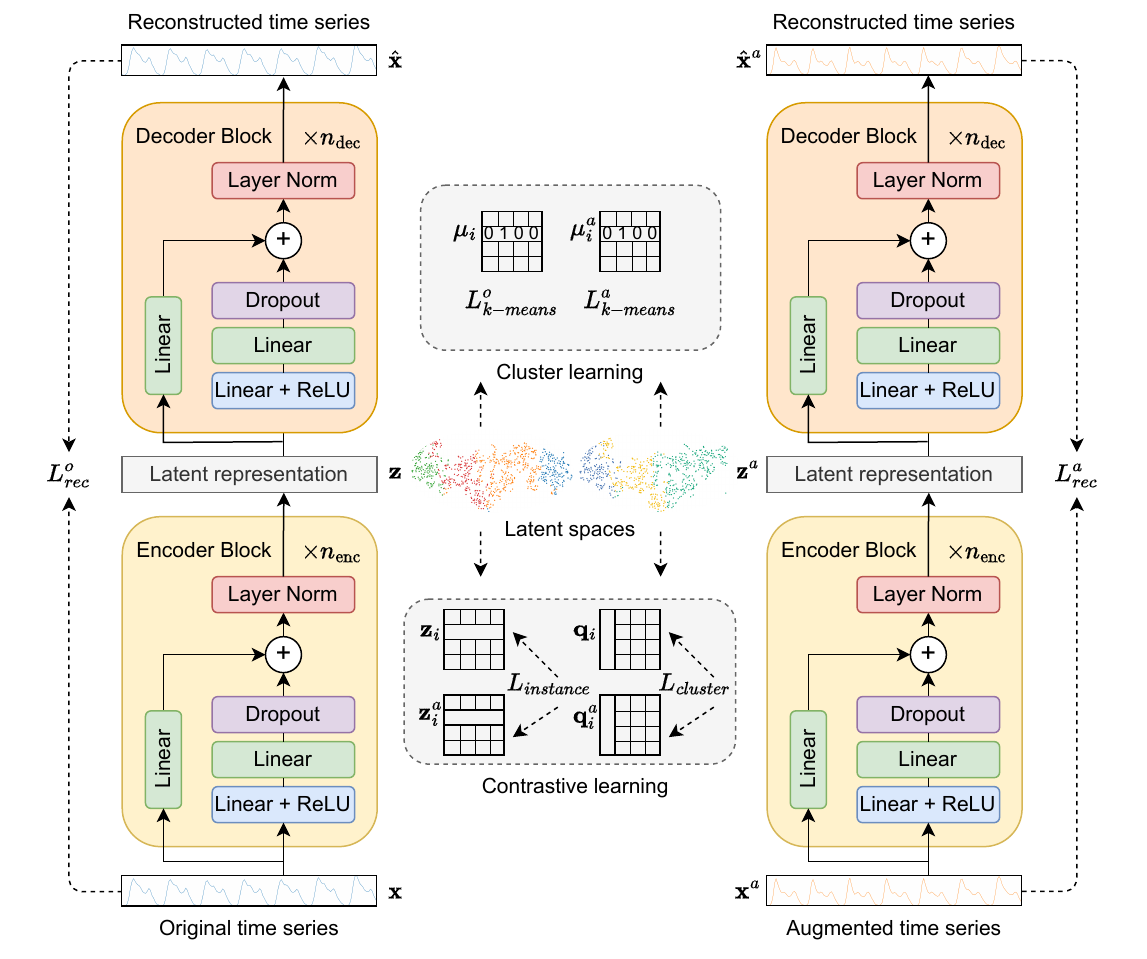}
  \caption{The overall architecture of LoSTer for long-sequence time series clustering (LSTC). It employs a two-view architecture with identical autoencoders for reconstructing original and augmented time series. Both the encoder and decoder are instantiated as a stack of dense residual blocks. The latent representations of data and centroids and the cluster assignments are learned end-to-end through \textit{k}-means and dual contrastive loss.}
  \label{fig:DenseAutoencoder}
\end{figure*}

Clustering is relevant for time series analysis in many applications as it helps discover temporal patterns from unlabeled data. However, the widespread collection of increasingly longer time series data in various sectors - like industry and finance - unveils the limits of current solutions. New challenges posed by long-sequence time series clustering (LSTC) emerge from the need to design temporal clustering algorithms that can effectively handle high dimensionality and achieve computational scalability when applied to large-scale datasets.

The \textit{k}-means algorithm is an easy and informative method to group data with high similarity within clusters, including time series \cite{aghabozorgi2015time,kobylin2020time}. However, the assumption of underlying spherical cluster distribution causes poor performance when time series exhibit non-linear dynamics, particularly if spanning over thousands of time steps. Therefore, the unsupervised learning of a feature space in which to perform clustering has become an active research field.

Rather than original data, feature-based methods cluster lower-dimensional representations that filter irrelevant information and reduce noise and outliers. However, these methods require domain expertise to handcraft such features properly. In light of this, deep learning has been proposed for shaping an \emph{ad hoc} latent space in which representation learning and clustering are simultaneously performed, leading to deep clustering. Deep Embedding for Clustering (DEC) \cite{xie2016unsupervised} emerged as the first joint optimization on static data (e.g. images): it defines a non-linear mapping parameterized by a neural network from the original to the latent space, where the clustering objective is optimized using gradient descent. The Improved Deep Embedding for Clustering (IDEC) \cite{guo2017improved} integrates a reconstruction term into DEC to preserve the local structure of the data, inspiring Madiraju et al. \cite{madiraju2018deep} to design Deep Temporal Clustering (DTC) as an extension of IDEC to time series thanks to a CNN-BiLSTM autoencoder performing spatiotemporal dimensionality reduction. The drawback, though, is DTC as well as DEC and IDEC refining clusters by minimizing the Kullback-Leibler divergence to an auxiliary target distribution based on the current soft cluster assignments, instead of the \textit{k}-means objective: this surrogate loss may be sub-optimal and favour overlapping compared to hard clustering \cite{gao2020deep}.

Ma et al. \cite{ma2019learning} proposed Deep Temporal Clustering Representation (DTCR), which integrates time series reconstruction and \textit{k}-means to generate cluster-specific temporal representations through a Dilated RNN \cite{chang2017dilated}. Zhong et al. \cite{zhong2023deep} further include contrastive learning in their Deep Temporal Contrastive Clustering (DTCC), but neither DTCR nor DTCC is optimizing the canonical \textit{k}-means objective due to the non-differentiability of the hard cluster assignment. Indeed, they seek to minimize a soft \textit{k}-means objective, which can be converted into a trace maximization problem via spectral relaxation \cite{zha2001spectral}. However, its implementation requires fixing the cluster assignments for a predefined number of iterations to avoid instability while updating data representations, causing inconsistent assignments in mini-batch stochastic gradient descent. LSTC may require many iterations if the dataset is large, eventually making this phenomenon more detrimental.

Therefore, we design Long-Sequence Temporal Concrete Clustering (LoSTer) resulting in the first end-to-end deep learning model solving the non-differentiability of the \textit{k}-means discrete cluster assignment of time series, by exploiting the Gumbel-softmax reparameterization trick as in Gao et al. \cite{jang2016categorical}.

Furthermore, LoSTer defines a simple and yet effective dense architecture to overcome 1) the limiting autoregressive single-layer RNN of DTCR and DTCC subject to error accumulation and slowing down training due to processing the observations in the sequence recursively and 2) the inadequacy of the point-wise attention mechanism in Transformers to capture temporal dependencies over long time series while also being hindered by quadratic computation cost and space requirements. The issues in Transformers, which are known findings from the long-term time series forecasting research \cite{zeng2023transformers}, are even exacerbated in LSTC because the aim is to cluster large batches of long time series as a whole rather than forecasting the future of each time series individually. Specifically, the high memory occupation caused by embedding each of the many temporal observations forces to set a smaller batch size: since deep clustering is performed batch-wise, smaller and therefore less representative batches cause a drop in performance. Conversely, linear models have shown in literature remarkable predictive capability in handling trends, seasonality and shifts in time series, while boasting higher efficiency \cite{zeng2023transformers,li2023revisiting}.

Experimental results on 17 long-sequence time series benchmark datasets show LoSTer outperforming state-of-the-art RNNs and Transformers in terms of clustering accuracy and training speed, thus eventually delivering a simple, fast to train and scalable method for LSTC. Moreover, superior performance on two more challenging large-scale retail datasets proves the effectiveness in real-world use cases. The main contributions of this work are summarized as follows:

\begin{enumerate}
    \item To the best of our knowledge in time series clustering, LoSTer is the first deep learning model that optimizes the \textit{k}-means objective without falling back to sub-optimal surrogates. Unlike DTCR and DTCC, cluster assignments are not fixed for a predefined number of iterations, avoiding inconsistent assignments.
    \item Moving from the solid findings of the long-range time series forecasting research, this work provides theoretical discussions on the poor capacity of state-of-the-art Transformers in performing LSTC.
    \item A novel dense autoencoder is designed specifically for LSTC, overcoming effectively the limitations of RNNs and Transformers.
    \item LoSTer boasts considerable training speedup to RNNs and Transformer-based competitors when applied to large-scale LSTC settings.
\end{enumerate}

\section{Related Works}
\textbf{Raw-based methods.} Time series clustering is divided into raw-based and feature-based methods. The traditional \textit{k}-means algorithm minimizes the square of intra-cluster Euclidean distance. When applied to time series, though, this metric is extremely restrictive because the time steps are matched point-wise. For this reason, raw-based methods are applied with a modified similarity measure. For example, \textit{k}-DBA \cite{petitjean2011global} uses Dynamic Time Warping (DTW) and performs clustering based on the aligned set of sequences. Similarly, K-Spectral Centroid (K-SC) \cite{yang2011patterns} clusters temporal sequences based on a similarity metric that is invariant to scaling and shifting. However, all raw-based methods are sensitive to noise and outliers because they involve every time step.

\textbf{Feature-Based Methods.} Instead of considering all time steps, feature-based methods reduce the data dimension and alleviate the impact of high-frequency noise, temporal gaps, and outliers. Examples include adaptive piece-wise constant approximation \cite{keogh2001locally}, nonnegative matrix factorization \cite{brockwell1991time}, and Independent Component Analysis (ICA) \cite{guo2008time}. However, they are two-stage approaches in which the dimensionality reduction and clustering are performed independently, causing the potential loss of temporal information.

\textbf{Deep Temporal Clustering.} The interest in time series deep clustering has grown due to the ability to learn a latent space that captures the properties of data without manual feature engineering. These models have also the benefit of jointly optimizing the feature extraction and clustering. DTC \cite{madiraju2018deep} uses an autoencoder and a clustering layer, trained by minimizing the Kullback-Leibler divergence between the predicted and target distribution. DTCR \cite{ma2019learning} incorporates the \textit{k}-means objective into the model via spectral relaxation. More recently, following the growing interest in contrastive learning for image clustering \cite{li2021contrastive,li2020prototypical,peng2022crafting}, DTCC \cite{zhong2023deep} has shown a novel dual contrastive learning approach yielding more informative discriminative features for time series clustering.

\textbf{Learning Representation from Time Series.} Regardless of their complicated self-attention scheme, a very simple linear model \cite{zeng2023transformers} has been able to outperform the state-of-the-art Transformers \cite{zhou2022fedformer,chen2021autoformer,zhou2021informer,liu2021pyraformer,li2019enhancing} on a variety of common long-term time series forecasting benchmarks \cite{zhou2021informer,misc_electricityloaddiagrams20112014_321,lai2018modeling}. This remarks that designing an encoder that learns an effective representation of long time series is still an open challenge. In this work, we design a new dense architecture for time series clustering partly inspired by the Time-Series Dense Encoder (TiDE) \cite{das2023long}, which achieves superior forecasting accuracy and faster computation when compared to a strong patch time series Transformer \cite{nie2022time}. Similarly, Preformer \cite{du2023preformer} and Pathformer \cite{chen2024pathformer} have recently proposed segment-wise and patch-level attention at multiple scales, respectively, to overcome the standard point-wise attention while also capturing different characteristics for various resolutions.

\section{Proposed Method}
We present LoSTer, a novel deep-learning method for long-sequence time series clustering. Compared to current state-of-the-art DTCR and DTCC, LoSTer is not limited to generating cluster-friendly representations of time series but is also able to predict the final clusters end-to-end, hence without the need of performing \textit{k}-means on the latent representations at the end of the training process. A new dense encoder-decoder architecture and the optimization are described in this section. The general scheme is illustrated in Fig. 1.

\subsection{Preliminaries}
Consider a set of $n$ time series  $D = \{ \mathbf{x}_1, ..., \mathbf{x}_n \}$ where each ordered sequence $\mathbf{x}_i$ contains $L$ time steps and is denoted as $\mathbf{x}_i = [x_{i,1}, ..., x_{i,L}]$. Given a positive integer $k \leq n$, we perform \textit{k}-means clustering to partition the $n$ instances in $D$ into $k$ clusters.

As this work uses a contrastive learning approach, we need to define an augmentation function $\textit{T}$. For each instance $\mathbf{x}_i$, we compute in advance its augmented version $\mathbf{x}_i^a = T(\mathbf{x}_i)$, where $T$ is the composition of rotation, permutation, time-warping as proposed by Um et al. \cite{um2017data} to perturb the temporal location of time series sensor data.

\subsection{Method Overview}
LoSTer uses the two-view architecture in which two identical autoencoders reconstruct the original and augmented time series while performing cluster learning in the corresponding latent spaces. The autoencoder is instantiated as a simple and efficient MLP-based architecture. The encoder is built as a stack of residual blocks mapping the input time series to a lower-dimensional feature vector. The decoder uses a similar stack to reconstruct the input sequence given the encoded representation. Meanwhile, exploiting the Gumbel-softmax reparameterization trick, the centroid representation matrix can be defined as a learnable parameter of the model, thus learned end-to-end without alternating with cluster assignments updates. The dual contrastive loss is enforced to exploit the two-view architecture.

\subsection{Dense Encoder-Decoder Architecture}
State-of-the-art solutions for time series clustering capture the temporal dynamics through the Dilated RNN, which extracts multi-scale features using dilated recurrent skip connections and exponentially increasing dilation. This multi-scale architecture improved the ability of recurrent models to learn time dependency. However, DTCR and DTCC still propose an autoregressive single-layer RNN as the decoder for reconstructing the input time series. This choice affects the reconstruction ability in the decoding phase because the autoregressive strategy is prone to error accumulation, particularly on longer sequences with intricate patterns.
To overcome this issue, we introduce a deep temporal clustering model featuring an MLP-based autoencoder architecture, whose decoder performs the reconstruction of the time series from the encoded representation using the multi-step direct strategy instead of autoregression.

\textbf{Residual Block.} The core component is a residual block with one hidden layer and a fully linear skip connection. Let $\mathbf{x}$ be the input vector to the module. The hidden layer with the ReLU activation function is followed by a linear transformation that maps the embedding $\mathbf{h}$ to the output representation $\mathbf{o}$. Dropout is enabled during training for regularization. Layer normalization is optionally applied at the end. In formulas:
\begin{equation}
\mathbf{h} = \text{ReLU}(\text{Linear}(\mathbf{x}))
\end{equation}
\begin{equation}
\mathbf{h'} = \text{Dropout}(\text{Linear}(\mathbf{h})) + \text{Linear}(\mathbf{x})
\end{equation}
\begin{equation}
\mathbf{o} = \text{LayerNorm}(\mathbf{h'})
\end{equation}
\textbf{Encoder.} The dense encoder \text{Enc} of LoSTer is a stack of $n_{\text{enc}}$ residual blocks mapping the input sequence $\mathbf{x}_i$ to a vector $\mathbf{z}_i$ in a $d$-dimensional latent space.
\begin{equation}
    \mathbf{z}_i = \text{Enc}(\mathbf{x}_i)
\end{equation}
\textbf{Decoder.} The dense decoder \text{Dec} is defined as a stack of $n_{\text{dec}}$ residual blocks with the same hidden size $d$ as the encoder, producing the reconstruction $\hat{x}_i$ of the input.
\begin{equation}
    \hat{x}_i = \text{Dec}(\mathbf{z}_i)
\end{equation}

\subsection{Deep Temporal Reconstruction with Parallel Views}
 We aim to learn two non-linear encoding transformations $f_{\text{enc}}: \mathbf{x}_i \rightarrow \mathbf{z}_i $ for the original view and $f_{\text{enc}}^a: \mathbf{x}_i^a \rightarrow \mathbf{z}_i^a $ for the augmented view, mapping the original and augmented sequence respectively into the latent space $\textit{Z}$ and $\textit{Z}^a$ $\subset \mathbb{R}^d$. Also, we define two non-linear decoding transformations $f_{\text{dec}}: \mathbf{z}_i \rightarrow \mathbf{\hat{x}}_i $ and $f_{\text{dec}}^a: \mathbf{z}_i^a \rightarrow \mathbf{\hat{x}}_i^a $ mapping the latent representations $\mathbf{z}_i$ and $\mathbf{z}_i^a$ back to the initial input space of each view. To capture the dynamics of time series, we instantiate the encoding and decoding functions $f_{\text{enc}}$ and  $f_{\text{dec}}$ as the encoder and decoder, respectively, of the proposed dense autoencoder described above. Similarly, the same neural networks are instantiated for the functions $f_{\text{enc}}^a$ and $f_{\text{dec}}^a$ in the augmented view. Note that the two autoencoders are identical but have different weights. To push the network to learn meaningful representations, we use mean squared error as the reconstruction loss for training the encoder-decoder composition in each view. Formally:
\begin{equation}
    L_{rec}^o = \frac{1}{n} \sum_{i=1}^n \lvert\lvert \mathbf{x}_i - \mathbf{\hat{x}}_i \rvert\rvert_2^2
\end{equation}
\begin{equation}
    L_{rec}^a = \frac{1}{n} \sum_{i=1}^n \lvert\lvert \mathbf{x}_i^a - \mathbf{\hat{x}}_i^a \rvert\rvert_2^2
\end{equation}
Therefore, the joint reconstruction loss is:
\begin{equation}
    L_{rec} = L_{rec}^o + L_{rec}^a
\end{equation}

\subsection{Differentiable Discrete Clustering}
Though needed for consistency, the reconstruction loss is not purposed to capture cluster-level information. To this end, we describe a cluster distribution learning method capable of optimizing the hard-cluster assignment of the \textit{k}-means objective directly, enabling end-to-end gradient-based learning, and therefore without resorting to a soft \textit{k}-means relaxation as opposed to existing works.

In deep \textit{k}-means, the latent representations in the embedding space $\textit{Z}$ are partitioned into $\textit{k}$ different clusters. The optimization problem is to find the assignment and set of $\textit{k}$ centroids that minimize the sum of squared distances between each $\mathbf{z}_i$ and its associated centroid. Formally:
\begin{equation}
\min_{\mathbf{Q},\mathbf{M}} \lvert\lvert \mathbf{Z} - \mathbf{Q}\mathbf{M} \rvert\rvert_F^2 \text{ s.t. $\lvert\lvert \mathbf{q}_i \rvert\rvert_1 = 1$} \text{, $\mathbf{Q} \in \{0, 1\}^{n \times k}$}
\end{equation}
where the row $\mathbf{q}_i$ of the binary matrix $\mathbf{Q}$ is a one-hot encoding representing the assignment of $\mathbf{z}_i$ to one cluster only. $\mathbf{M} \in \mathbb{R}^{k \times d}$ is the centroid matrix, in which rows $\bm{\mu}_j$ are the centroid representations in the latent space.

The \textit{k}-means algorithm alternates the cluster assignments and cluster centers optimization until convergence. For deep \textit{k}-means clustering, though, backpropagation is not possible as the discrete cluster assignment is not differentiable. Therefore, we exploit the Gumbel-softmax reparameterization trick, which uses a stochastic hard assignment in the forward pass while allowing gradient backpropagation through the soft assignment. This enables the end-to-end training of the neural network.
\begin{equation}
    p(C_{i,j} | \mathbf{x}_i) = \frac{\exp{\{ -\sigma^{-2} \lvert\lvert \mathbf{z}_i - \bm{\mu}_j \rvert\rvert_2^2 \}}}{\sum_{c=1}^k \exp \{ -\sigma^{-2} \lvert\lvert \mathbf{z}_i - \bm{\mu}_c \rvert\rvert_2^2 \}}
    \label{rbf}
\end{equation}
As shown in Eq. \ref{rbf}, the cluster assignment probabilities are based on the Euclidean distance and modeled using the normalized radial basis function (RBF). Specifically, $p(C_{i,j} | \mathbf{x}_i)$ is the probability of the event that $\mathbf{x}_i$ is assigned to cluster $j$, whose centroid is $\bm{\mu}_j$, considering all the centroid distances of the corresponding representation $\mathbf{z}_i$. At this point, we use the reparameterization trick in Eq. \ref{gumbel} to draw samples $\mathbf{q}_i$ from a categorical distribution with class probabilities $p(C_{i,j} | \mathbf{x}_i)$ for $j \in \{1, ..., k\}$, while still being able to perform backpropagation through the sampling process.
\begin{equation}
    q_{i,j} = \frac{\exp{\{ (\log({p(C_{i,j} | \mathbf{x}_i)}) + g_j ) / \tau \}}}{\sum_{c=1}^k \exp{\{ (\log({p(C_{i,c} | \mathbf{x}_i)}) + g_c) / \tau \}} }
    \label{gumbel}
\end{equation}
Here $q_{i,j}$ is the \emph{j}-th entry of $\mathbf{q}_i$, which stands as an approximation of the \textit{k}-dimensional one-hot vector representing the hard cluster assignment of $\mathbf{x}_i$. $g_1, ... g_k$ are i.i.d. samples drawn from Gumbel(0,1), and $\tau$ is a hyperparameter controlling the entropy of the Gumbel-softmax distribution: the softmax distribution converges to a categorical distribution when $\tau \rightarrow 0$. As we decrease $\tau$ during training, the probabilistic hard assignment $\mathbf{q}_i$ can be discretized by rounding the largest component to 1 and all others to 0. In this way, the model outputs the true discrete sample $\mathbf{\tilde{q}}_i$ distributed according to $p(C_i | \mathbf{x}_i)$. Now, we can define the \textit{k}-means loss function for the original and augmented view.
\begin{equation}
    L_{k-means}^o = \frac{1}{n} \sum_{i=1}^n \lvert\lvert \mathbf{z}_i - \mathbf{\tilde{q}}_i\mathbf{M} \rvert\rvert_2^2
    \label{kmeans_o}
\end{equation}
\begin{equation}
    L_{k-means}^a = \frac{1}{n} \sum_{i=1}^n \lvert\lvert \mathbf{z}_i^a - \mathbf{\tilde{q}}_i^a\mathbf{M}_a \rvert\rvert_2^2
    \label{kmeans_a}
\end{equation}
\begin{equation}
    L_{k-means} = \frac{1}{2} (L_{k-means}^o + L_{k-means}^a)
\end{equation}
The \textit{k}-means formulation in Eq. \ref{kmeans_o} and \ref{kmeans_a} based on one-hot encoded categorical variables enables the model to control and learn the data representation and discrete assignment simultaneously, while also learning explicitly the centroid representation since $\mathbf{M}$ and $\mathbf{M}_a$ are defined indeed as parameters of the model.

\subsection{Contrastive Learning}
The information from the original and the augmented view are bridged together via dual contrastive learning that captures the descriptive patterns at both the instance and cluster levels.

\textbf{Instance Contrastive Loss.} The instance contrastive loss seeks to cluster the latent representation of a time series $\mathbf{z}_i$ and its augmented version $\mathbf{z}_i^a$, while minimizing the similarity to other time series representations.
\begin{equation} \label{instance_contrastive_loss}
    \resizebox{0.48\textwidth}{!}{$l_{z_i} = -\log{\frac{\exp{\{\text{sim}(\mathbf{z}_i, \mathbf{z}_i^a) / \tau_I}\}}{\sum_{j=1}^n [\exp{\{ \text{sim}(\mathbf{z}_i, \mathbf{z}_j) / \tau_I \}} + \mathbbm{1}_{[i \neq j]} \exp{\{ \text{sim}(\mathbf{z}_i, \mathbf{z}_j^a) / \tau_I \}}]}}$}
\end{equation}
$\text{sim}(\mathbf{z}_i,\mathbf{z}_i^a)$ is the dot product between the L2-normalized vectors of $\mathbf{z}_i$, $\mathbf{z}_i^a$ and $\tau_I$ is the temperature hyperparameter balancing the influence of negative pairs versus one positive pair. By defining the augmented version $l_{z_i^a}$ similarly, the overall instance contrastive loss is:
\begin{equation}
    L_{instance} = \frac{1}{2n} \sum_{i=1}^n (l_{z_i} + l_{z_i^a})
\end{equation}
\textbf{Cluster Contrastive Loss.} LoSTer uses a cluster contrastive loss acting on the probabilistic hard assignments in Eq. \ref{gumbel} to favour the cluster distribution of the two views. Considering that a row in $\mathbf{Q}$ is the probability distribution of the class assignment of an instance over \textit{k} centroids, we now refer to the \emph{i}-th column as $\mathbf{q}_i$, which can be interpreted as the representation of the \emph{i}-th cluster. Therefore, we aim to enforce higher similarity between the corresponding columns $\mathbf{q}_i$ and $\mathbf{q}_i^a$ in the views, while minimizing it for all other pairs, which are considered negative samples. $\tau_C$ is the cluster-wise temperature hyperparameter.
\begin{equation} \label{cluster_contrastive_loss}
    \resizebox{0.48\textwidth}{!}{$l_{q_i} = -\log{\frac{\exp{\{\text{sim}(\mathbf{q}_i, \mathbf{q}_i^a) / \tau_C}\}}{\sum_{j=1}^n [\exp{\{ \text{sim}(\mathbf{q}_i, \mathbf{q}_j) / \tau_C \}} + \mathbbm{1}_{[i \neq j]} \exp{\{ \text{sim}(\mathbf{q}_i, \mathbf{q}_j^a) / \tau_C \}}]}}$}
\end{equation}
To avoid a degenerate solution resulting in one large cluster and multiple isolated small ones, we need to incorporate the entropy of the cluster-assignment probability.
\begin{equation} \label{cluster_entropy}
    H = -\sum_{i=1}^k p(\mathbf{q}_i) \log p(\mathbf{q}_i) + p(\mathbf{q}_i^a) \log p(\mathbf{q}_i^a)
\end{equation}
Here, $p(\mathbf{q}_i)$ is the average assignment probability for the \emph{i}-th cluster, i.e. formally $p(\mathbf{q}_i) = \frac{1}{n} \sum_{j=1}^n \mathbf{q}_{j,i}$. This is the cluster contrastive loss.
\begin{equation}
    L_{cluster} = \frac{1}{2k} \sum_{i=1}^k (l_{q_i} + l_{q_i^a}) - H
\end{equation}

\subsection{Overall Loss Function}
\begin{algorithm}
\caption{LoSTer Training Method}\label{alg:cap}
\begin{algorithmic}
\STATE {\bfseries Input:} Dataset: $D$, Number of clusters: $k$, Augmentation function: \textit{T}, Temperature: $\tau$, Learning rate: $\alpha$
\STATE {\bfseries Output:} Cluster assignment matrix: $\mathbf{Q}$
\STATE For each $\mathbf{x}_i \in D$, generate $\mathbf{x}_i^a$ applying \textit{T}
\STATE Init model parameters $\Phi$, $\Phi^a$
\WHILE{not converged}
\STATE $\Phi \gets \Phi - \alpha \nabla_{\Phi} L_{rec}^o$
\ENDWHILE
\WHILE{not converged}
\STATE $\Phi^a \gets \Phi^a - \alpha \nabla_{\Phi^a} L_{rec}^a$
\ENDWHILE
\STATE Init centroid representations $\mathbf{M}$, $\mathbf{M}^a$ with \textit{k}-means++
\WHILE{stop criterion is not verified}
\STATE Adjust $\tau$
\STATE $\Phi \gets \Phi - \alpha \nabla_{\Phi} L_{LoSTer}$
\STATE $\Phi^a \gets \Phi^a - \alpha \nabla_{\Phi^a} L_{LoSTer}$
\STATE $\mathbf{M} \gets \mathbf{M} - \alpha \nabla_{M} L_{LoSTer}$
\STATE $\mathbf{M}^a \gets \mathbf{M}^a - \alpha \nabla_{M^a} L_{LoSTer}$
\ENDWHILE
\STATE Get $\mathbf{Q}$ applying $\arg \max$ of $p(C_i | \mathbf{x}_i)$
\end{algorithmic}
\end{algorithm}
Finally, we define the overall training loss $L_{LoSTer}$ and describe the details of the training algorithm.
\begin{equation}
    L_{LoSTer} = L_{rec} + L_{k-means} + L_{instance} + L_{cluster}
\end{equation}
In practice, the autoencoders for the original and augmented time series are trained separately using only the reconstruction loss, then used as feature extractors for the initialization of the centroids in their respective latent space, before jointly being optimized under the full objective $L_{LoSTer}$ to produce the final cluster assignment. Note that we start training with a high temperature $\tau$ and perform linear annealing to decrease it towards zero as training progresses gradually. As opposed to the strategy of DTCR and DTCC, LoSTer jointly updates the data representation and cluster assignment at each step without resorting to alternate optimization. Clustering results are also obtained end-to-end instead of applying \textit{k}-means to the learned representations when training is over.

\section{Experiments}
\begin{table}[!htb]
  \caption{Details of the benchmark UCR time series datasets.}
  \label{ucr-datasets}
  \vskip 0.1in
  \begin{center}
  \begin{small}
  \begin{sc}
  \resizebox{\columnwidth}{!}{
  \begin{tabular}{llll}
    \toprule
    Dataset & Train/Test & Length & Classes\\
    \midrule
    CinCECGTorso & 40/1380 & 1639 & 4\\
    NonInvasiveFetalECGThorax1 & 1800/1965 & 750 & 42\\
    NonInvasiveFetalECGThorax2 & 1800/1965 & 750 & 42\\
    StarLightCurves & 1000/8236 & 1024 & 3\\
    UWaveGestureLibraryX & 896/3582 & 315 & 8\\
    UWaveGestureLibraryY & 896/3582 & 315 & 8\\
    MixedShapesRegularTrain & 500/2425 & 1024 & 5\\
    MixedShapesSmallTrain & 100/2425 & 1024 & 5\\
    EOGVerticalSignal & 362/362 & 1250 & 12\\
    SemgHandMovementCh2 & 450/450 & 1500 & 6\\
    ECG5000 & 500/4500 & 140 & 5\\
    OSULeaf & 200/242 & 427 & 6\\
    Symbols & 25/995 & 398 & 6\\
    MiddlePhalanxOutlineAgeGroup & 400/154 & 80 & 3\\
    ProximalPhalanxOutlineAgeGroup & 400/205 & 80 & 3\\
    ProximalPhalanxTW & 400/205 & 80 & 6\\
    SyntheticControl & 300/300 & 60 & 6\\
    \bottomrule
  \end{tabular}
  }
  \end{sc}
  \end{small}
  \end{center}
  \vskip -0.1in
\end{table}

\begin{table*}[!tb]
  \caption{Rand Index (RI) for UCR datasets. A higher RI is better and the best results are highlighted.}
  \label{ri-table}
  \vskip 0.1in
  \begin{center}
  \begin{small}
  \begin{sc}
  \begin{tabular}{llllllll}
    \toprule
    Dataset & LoSTer & IDEC & DNFCS & CKM & DTC & DTCR & DTCC\\
    \midrule
    CinCECGTorso & \textbf{0.6906} & 0.6826 & 0.6799 & 0.6785 & 0.5973 & 0.6072 & 0.6339\\
    NonInvasiveFetalECGThorax1 & \textbf{0.9737} & 0.9671 & 0.9668 & 0.9668 & 0.9621 & 0.9548 & 0.9582\\
    NonInvasiveFetalECGThorax2 & \textbf{0.9781} & 0.9728 & 0.9726 & 0.9727 & 0.972 & 0.9559 & 0.9616\\
    StarLightCurves & \textbf{0.7661} & 0.759 & 0.7582 & 0.761 & 0.7628 & 0.7622 & 0.6833\\
    UWaveGestureLibraryX & \textbf{0.8602} & 0.853 & 0.8529 & 0.8522 & 0.8248 & 0.704 & 0.8154\\
    UWaveGestureLibraryY & \textbf{0.8483} & 0.8446 & 0.8442 & 0.8432 & 0.8163 & 0.7282 & 0.8139\\
    MixedShapesRegularTrain & \textbf{0.8259} & 0.8139 & 0.8144 & 0.812 & 0.7483 & 0.6028 & 0.5926\\
    MixedShapesSmallTrain & \textbf{0.8257} & 0.8174 & 0.8158 & 0.8165 & 0.6131 & 0.6219 & 0.6490\\
    EOGVerticalSignal & \textbf{0.8669} & 0.8474 & 0.8456 & 0.836 & 0.815 & 0.844 & 0.8251\\
    SemgHandMovementCh2 & \textbf{0.7519} & 0.7509 & 0.7499 & 0.7453 & 0.6813 & 0.6302 & 0.7207\\
    ECG5000 & \textbf{0.7142} & 0.7083 & 0.7088 & 0.7123 & 0.7003 & 0.6832 & 0.6727\\
    OSULeaf & \textbf{0.7583} & 0.7415 & 0.7416 & 0.7421 & 0.7287 & 0.705 & 0.7063\\
    Symbols & \textbf{0.8995} & 0.8931 & 0.8931 & 0.8934 & 0.8779 & 0.7972 & 0.8988\\
    MiddlePhalanxOutlineAgeGroup & \textbf{0.7355} & 0.7288 & 0.734 & 0.7287 & 0.7095 & 0.7263 & 0.7305\\
    ProximalPhalanxOutlineAgeGroup & \textbf{0.8} & 0.7795 & 0.784 & 0.7797 & 0.7907 & 0.7929 & 0.794\\
    ProximalPhalanxTW & \textbf{0.8748} & 0.8092 & 0.7894 & 0.8038 & 0.7892 & 0.7846 & 0.8197\\
    SyntheticControl & 0.8698 & 0.8474 & 0.8474 & 0.8476 & \textbf{0.8763} & 0.8389 & 0.8561\\
    \midrule
    Average RI & \textbf{0.8259} & 0.8127 & 0.8117 & 0.8113 & 0.7803 & 0.7494 & 0.7725\\
    \% difference &  & -1.32\% & -1.42\% & -1.46\% & -4.56\% & -7.65\% & -5.34\%\\
    p-value ($\alpha=0.05$) &  & 3e-4 & 3e-4 & 3e-4 & 3e-4 & 3e-4 & 3e-4\\
    \bottomrule
  \end{tabular}
  \end{sc}
  \end{small}
  \end{center}
  \vskip -0.1in
\end{table*}

\begin{table*}[!tb]
  \caption{Normalized Mutual Index (NMI) for UCR datasets. A higher NMI is better and the best results are highlighted.}
  \label{nmi-table}
  \vskip 0.1in
  \begin{center}
  \begin{small}
  \begin{sc}
  \begin{tabular}{llllllll}
    \toprule
    Dataset & LoSTer & IDEC & DNFCS & CKM & DTC & DTCR & DTCC\\
    \midrule
    CinCECGTorso & \textbf{0.2812} & 0.2599 & 0.2646 & 0.2757 & 0.0379 & 0.0462 & 0.0548\\
    NonInvasiveFetalECGThorax1 & \textbf{0.7399} & 0.6755 & 0.6764 & 0.6754 & 0.6129 & 0.6004 & 0.588\\
    NonInvasiveFetalECGThorax2 & \textbf{0.8019} & 0.7577 & 0.7608 & 0.7579 & 0.7474 & 0.6271 & 0.6577\\
    StarLightCurves & \textbf{0.6024} & 0.5567 & 0.5504 & 0.5554 & 0.5317 & 0.488 & 0.3659\\
    UWaveGestureLibraryX & \textbf{0.4516} & 0.4384 & 0.439 & 0.437 & 0.3366 & 0.1351 & 0.2663\\
    UWaveGestureLibraryY & \textbf{0.4388} & 0.4199 & 0.4195 & 0.4183 & 0.3811 & 0.189 & 0.189\\
    MixedShapesRegularTrain & \textbf{0.5147} & 0.4841 & 0.4844 & 0.483 & 0.321 & 0.0861 & 0.0895\\
    MixedShapesSmallTrain & \textbf{0.5122} & 0.5034 & 0.5011 & 0.512 & 0.1079 & 0.0845 & 0.0941\\
    EOGVerticalSignal & \textbf{0.3619} & 0.3065 & 0.3065 & 0.3109 & 0.1583 & 0.1655 & 0.1655\\
    SemgHandMovementCh2 & \textbf{0.2508} & 0.2302 & 0.2276 & 0.2153 & 0.1235 & 0.1136 & 0.1032\\
    ECG5000 & \textbf{0.4781} & 0.4699 & 0.4699 & 0.4709 & 0.4566 & 0.3557 & 0.36\\
    OSULeaf & \textbf{0.2286} & 0.2091 & 0.2096 & 0.2139 & 0.1356 & 0.0871 & 0.0794\\
    Symbols & \textbf{0.7979} & 0.764 & 0.7639 & 0.7655 & 0.7531 & 0.5816 & 0.7595\\
    MiddlePhalanxOutlineAgeGroup & 0.392 & 0.39 & \textbf{0.3991} & 0.3891 & 0.2932 & 0.3451 & 0.3569\\
    ProximalPhalanxOutlineAgeGroup & \textbf{0.5298} & 0.511 & 0.5231 & 0.512 & 0.494 & 0.5047 & 0.5089\\
    ProximalPhalanxTW & \textbf{0.6395} & 0.5266 & 0.513 & 0.5195 & 0.5218 & 0.5405 & 0.5656\\
    SyntheticControl & \textbf{0.7862} & 0.7119 & 0.7119 & 0.7142 & 0.724 & 0.5428 & 0.5878\\
    \midrule
    Average NMI & \textbf{0.5181} & 0.4832 & 0.4836 & 0.4839 & 0.3963 & 0.3231 & 0.3407\\
    \% difference &  & -3.49\% & -3.45\% & -3.42\% & -12.18\% & -19.5\% & -17.74\%\\
    p-value ($\alpha=0.05$) &  & 3e-4 & 3e-4 & 3e-4 & 3e-4 & 3e-4 & 3e-4\\
    \bottomrule
  \end{tabular}
  \end{sc}
  \end{small}
  \end{center}
  \vskip -0.1in
\end{table*}

Experiments are conducted on 17 long-sequence time series benchmark datasets from the UCR archive \cite{dau2019ucr} and two large-scale datasets from real-world applications \cite{makridakis2022m5,huy2023store}. The clustering performance is evaluated using two metrics: the Rand Index (RI) and the Normalized Mutual Information (NMI). For a fair comparison, we report the average scores over 3 independent runs of each experiment for all datasets.

We set 3 as the number of residual blocks $n_{\text{enc}}$ and $n_{\text{dec}}$ for all experiments with LoSTer, set $d=256$ as the dimension of the hidden representation and dropout at 0.1. The temperature hyperparameters $\tau$, $\tau_I$, and $\tau_C$ are set respectively to 10, 1, and 1. The Adam optimizer is initialized with the learning rate of $10^{-3}$ for pretraining the model under the reconstruction loss for 50 epochs per view, before the \textit{k}-means++ initialization for centroids in the latent space; then, we start training the model through stochastic gradient descent with a learning rate of $10^{-2}$ under the complete cost function and decaying by 0.1 every 5 epochs. The temperature $\tau$ is annealed according to the schedule $\tau = \max(\tau \cdot \beta^{epoch}, 0.01)$, where $\beta = 0.65$. As done by \cite{xie2016unsupervised,guo2017improved}, the training procedure is stopped when less than 0.1\% of samples change cluster assignment between two consecutive epochs, up to a maximum of 100 epochs.

We have found this setup working generally well for LoSTer. Specific tuning may further improve performance, although it would be time-consuming to perform on each dataset. For competing models, we use the same configuration as in their corresponding papers. The batch size is 128.

\subsection{Time Series Datasets}
The details of used datasets from the UCR Time Series Classification Archive \cite{dau2019ucr} are reported in Table \ref{ucr-datasets}. They have a default train/test split. However, since we are using these datasets as unlabeled data, we combine the train and test partitions like in Madiraju et al. \cite{madiraju2018deep}. Each time series is scaled individually by subtracting the mean and dividing by the standard deviation.

We have considered for LSTC 1) datasets with thousands of samples and time steps (e.g. CinCECGTorso, StarLightCurves, MixedShapesRegularTrain, MixedShapesSmallTrain), 2) large datasets with less than 1000 time steps (e.g. NonInvasiveFetalECGThorax1, NonInvasiveFetalECGThorax2, UWaveGestureLibraryX, UWaveGestureLibraryY, ECG5000, Symbols), 3) smaller datasets with hundreds of long sequences (e.g. SemgHandMovementCh2), 4) datasets with hundreds of instances and shorter sequences (e.g. OSULeaf, MiddlePhalanxOutlineAgeGroup, ProximalPhalanxOutlineAgeGroup, ProximalPhalanxTW, SyntheticControl).

\subsection{Evaluation metrics}
The Rand Index (RI) is a measure of the similarity between two sets of clusters. It measures the proportion of pairs of data samples correctly assigned to the same cluster ($TP$) or correctly assigned to different clusters ($TN$).
\begin{equation}
    RI = \frac{TP+TN}{n(n-1)/2}
\end{equation}
The Normalized Mutual Information (NMI) is a measure commonly used to evaluate the similarity between two clustering results. The mutual information between clusters $G_i$ and $A_j$ is divided by the squared root of the product of their entropies. $N$ is the total number of time series and $N_{ij} = \lvert G_i \cap A_j \rvert$.
\begin{equation}
    NMI = \frac{\sum_{i=1}^M \sum_{j=1}^M N_{ij} \log{(\frac{N \cdot N_{ij}}{\lvert G_i \rvert \lvert A_j \rvert})}}{\sqrt{(\sum_{i=1}^M \lvert G_i \rvert \log{\frac{\lvert G_i \rvert}{N}})(\sum_{j=1}^M \lvert A_j \rvert \log{\frac{\lvert A_j \rvert}{N}})}}
\end{equation}
In both these metrics, values close to 1 indicate well-performing clustering.

\subsection{Comparison with Deep Temporal Clustering Models}
LoSTer is compared to IDEC, Deep normalized fuzzy compactness and separation (DNFCS) \cite{feng2020deep}, Concrete \textit{k}-means (CKM) \cite{gao2020deep}, DTC, DTCR, and DTCC. However, we replace the CNN-BiLSTM in DTC with the RNN autoencoder used by DTCR and DTCC because of the need to adapt the number of filters, kernel and pooling size for each dataset. This choice, though, lets us observe how the same RNN architecture performs driven by different optimization strategies.

As shown in Table \ref{ri-table}, LoSTer achieves the highest average RI 0.8259 and the number of best results 16. It marks +7.65\% and +5.34\% differences over DTCR and DTCC, showing the limitations of RNN-based models optimized through soft \textit{k}-means for long-sequence time series clustering. The RI score by the modified DTC is 4.56\% lower on average than LoSTer, but still better than DTCR and DTCC. The MLP-based IDEC, DNFCS, and CKM overcome them with margin, then being outperformed in turn by LoSTer. Similarly, in Table \ref{nmi-table}, LoSTer achieves the highest average NMI of 0.5181 with 16 best results, higher than the scores of DTCR and DTCC by 19.5\% and 17.74\% respectively while the average gain in terms of NMI to IDEC, DNCFS, CKM is around +3.45\%. Reading the NMI scores row-wise in Table \ref{nmi-table}, datasets with very low NMI for DTCR and DTCC but higher for LoSTer emerge such as CinCECGTorso, MixedShapesRegularTrain, MixedShapesSmallTrain, and OsuLeaf.

The Wilcoxon signed rank test is conducted to measure the significance of the difference by comparing each method against LoSTer. As reported at the bottom of Table \ref{ri-table}, LoSTer is significantly better than competitors at the 0.05 p-value level.

\subsection{Comparison with Transformers}
\begin{table*}[!tb]
  \caption{Rand Index (RI) for UCR datasets. A higher RI is better and the best results are highlighted.}
  \label{ri-table-transformers}
  \vskip 0.1in
  \begin{center}
  \begin{small}
  \begin{sc}
  \begin{tabular}{lllll}
    \toprule
    Dataset & LoSTer & Preformer & Pathformer & iTransformer\\
    \midrule
    CinCECGTorso & \textbf{0.6906} & 0.5547 & 0.6199 & 0.6203\\
    NonInvasiveFetalECGThorax1 & \textbf{0.9737} & 0.7387 & 0.8281 & 0.8029\\
    NonInvasiveFetalECGThorax2 & \textbf{0.9781} & 0.9182 &  0.8397 & 0.7524\\
    StarLightCurves & 0.7661 & 0.5779 & 0.6441 & \textbf{0.7695}\\
    UWaveGestureLibraryX & \textbf{0.8602} & 0.6941 & 0.8019 & 0.7841\\
    UWaveGestureLibraryY & \textbf{0.8483} & 0.6545 & 0.8266 & 0.7949\\
    MixedShapesRegularTrain & \textbf{0.8259} & 0.6219 & 0.5109 & 0.717\\
    MixedShapesSmallTrain & \textbf{0.8257} & 0.5568 & 0.3906 & 0.7389\\
    EOGVerticalSignal & \textbf{0.8669} & 0.7105 & 0.6287 & 0.5993\\
    ECG5000 & 0.7142 & 0.6691 & 0.4716 & \textbf{0.7936}\\
    OSULeaf & \textbf{0.7583} & 0.4284 & 0.6438 & 0.4257\\
    Symbols & \textbf{0.8995} & 0.7021 & 0.6962 & 0.6897\\
    \midrule
    Average RI & \textbf{0.8339} & 0.6522 & 0.6585 & 0.7074\\
    \% difference &  & -18.17\% & -17.54\% & -12.65\%\\
    p-value ($\alpha=0.05$) &  & 2.22e-3 & 2.22e-3 & 9.6e-3\\
    \bottomrule
  \end{tabular}
  \end{sc}
  \end{small}
  \end{center}
  \vskip -0.1in
\end{table*}

\begin{table*}[!tb]
  \caption{Normalized Mutual Index (NMI) for UCR datasets. A higher NMI is better and the best results are highlighted.}
  \label{nmi-table-transformers}
  \vskip 0.1in
  \begin{center}
  \begin{small}
  \begin{sc}
  \begin{tabular}{lllll}
    \toprule
    Dataset & LoSTer & Preformer & Pathformer & iTransformer\\
    \midrule
    CinCECGTorso & \textbf{0.2812} & 0.0179 & 0.1608 & 0.0961\\
    NonInvasiveFetalECGThorax1 & \textbf{0.7399} & 0.245 & 0.5077 & 0.428\\
    NonInvasiveFetalECGThorax2 & \textbf{0.8019} & 0.4122 & 0.5649 & 0.4118\\
    StarLightCurves & 0.6024 & 0.1178 & 0.3515 & \textbf{0.6349}\\
    UWaveGestureLibraryX & \textbf{0.4516} & 0.1461 & 0.3071 & 0.3186\\
    UWaveGestureLibraryY & \textbf{0.4388} & 0.1884 & 0.3707 & 0.3617\\
    MixedShapesRegularTrain & \textbf{0.5147} & 0.0703 & 0.1655 & 0.319\\
    MixedShapesSmallTrain & \textbf{0.5122} & 0.0432 & 0.106 & 0.3979\\
    EOGVerticalSignal & \textbf{0.3619} & 0.1604 & 0.2318 & 0.1824\\
    ECG5000 & 0.4781 & 0.3175 & 0.0096 & \textbf{0.4929}\\
    OSULeaf & \textbf{0.2286} & 0.0457 & 0.1621 & 0.0383\\
    Symbols & \textbf{0.7979} & 0.5067 & 0.4762 & 0.5121\\
    \midrule
    Average NMI & \textbf{0.5174} & 0.1893 & 0.2845 & 0.2393\\
    \% difference &  & -32.81\% & -23.29\% & -27.81\%\\
    p-value ($\alpha=0.05$) &  & 2.22e-3 & 2.22e-3 & 4.8e-3\\
    \bottomrule
  \end{tabular}
  \end{sc}
  \end{small}
  \end{center}
  \vskip -0.1in
\end{table*}

The general-purpose architecture of the Transformer has garnered significant interest across various research domains like Natural Language Processing (NLP) \cite{vaswani2017attention} and Computer Vision \cite{dosovitskiy2020image} thanks to the attention mechanism's capability to effectively capture long-range dependencies in sequential data, suggesting it as a promising avenue for LSTC. However, the point-wise attention does not suit the individual time points because they do not deliver any semantic information as opposed to words in sentences, nor does the attention keep the sequential order of the time series because it is invariant to permutations. Moreover, the quadratic cost of computing the attention map and the memory requirements for embedding each temporal token of the sequence cause a computational bottleneck that prevents Transformers from dealing with very long time series.

At this stage, to extend the current validation by assessing the impact of these issues, we keep the same optimization strategy by LoSTer while replacing its dense architecture with the encoder of various recent Transformer long-term forecasters. Preformer (ICASSP 2023) \cite{du2023preformer} and Pathformer (ICLR 2024) \cite{chen2024pathformer} propose segment-wise and patch-level attention at multiple scales, respectively, to overcome the point-wise attention. Despite their complex structure, they are ruled by the dense autoencoder of LoSTer in terms of RI and NMI. Specifically, as shown in Table \ref{ri-table-transformers}, LoSTer marks the highest average RI 0.8339 overcoming Preformer (0.6522, i.e. +18.17\%) and Pathformer (0.6585, i.e. +17.54\%) for all datasets. Similarly, LoSTCO is the best model of the three also in terms of NMI as shown in Table \ref{nmi-table-transformers}, where its 0.5174 is superior to Preformer's 0.1893 (+32.81\%) and Pathformer's 0.2845 (+23.29\%). Massive improvements in NMI emerge in particular for bigger datasets like MixedShapesRegularTrain and MixedShapesSmallTrain, which comprise more than 2500 time series made of 1024 time steps. We underscore that the batch size for Preformer has been reduced from 128 to 16 due to exceeding memory usage. These results prove that extracting insightful long-term dependencies for LSTC is still challenging for the most sophisticated attention-based models.

LoSTCO also overcomes the state-of-the-art iTransformer (ICLR 2024) \cite{liu2024itransformer}, which proposes a change of perspective as it embeds the multivariate time series over the feature axis (rather than the temporal axis as usual) to pass over the intrinsic unalignment between potential delayed events or distinct physical measurements in the real world: consequently, embedding the distinct variate information at each time step actually worsens this unalignment and results in meaningless attention maps. Therefore, their model works out the time series by inverting the dimensions without modifying the Transformer architecture. Note that under univariate scenarios  - like in clustering use cases - iTransformer reduces to nothing more than a stackable linear predictor because there is only one variate to embed (attention degradation). This finding reveals an urgent gap in research about designing Transformers that can effectively exploit the temporal dependency in univariate time series and scale better than linear predictors; on the other hand, for multivariate data, the inversion in iTransformer adds a strong motivation for the prevalence of the channel-independence setup over channel-mixing both for linear models and Transformer predictors \cite{nie2022time}.

Regarding the validation, LoSTer outperforms iTransformer in 10 cases out of 12 in terms of RI (Table \ref{ri-table-transformers}) and NMI (Table \ref{nmi-table-transformers}), being slightly surpassed only when considering the StartLightCurves and EOGVerticalSignal datasets. However, iTransformer severely underperforms in terms of NMI for CinCECGTorso and OSULeaf benchmarks, eventually resulting in the global average RI 0.8339 by LoSTer being superior to iTransformer's 0.7074 (+12.65\%) and the average NMI 0.5174 to 0.2393 (+27.81\%).

The significant differences reported in Tables \ref{ri-table-transformers} and \ref{nmi-table-transformers} at the bottom are confirmed by performing the Wilcoxon signed rank test at the 0.05 p-value level.

\subsection{Ablation Study}
\begin{table*}[!tb]
  \caption{Rand Index (RI) ablation studies for UCR datasets. A higher RI is better and the best results are highlighted.}
  \label{ri-ablation-table}
  \vskip 0.1in
  \begin{center}
  \begin{small}
  \begin{sc}
  \begin{tabular}{lllllll}
    \toprule
    Dataset & LoSTer & MLP-based & RNN-based & Soft \textit{k}-means & KL & w/o CL\\
    \midrule
    CinCECGTorso & \textbf{0.6906} & 0.6751 & 0.6365 & 0.64 & 0.6853 & 0.6865\\
    NonInvasiveFetalECGThorax1 & 0.9737 & 0.9685 & 0.9636 & 0.9683 & \textbf{0.9741} & 0.9725\\
    NonInvasiveFetalECGThorax2 & \textbf{0.9781} & 0.9713 & 0.9701 & 0.9708 & 0.9771 & 0.9763\\
    StarLightCurves & 0.7661 & 0.7618 & 0.7683 & 0.7486 & 0.7662 & \textbf{0.7695}\\
    UWaveGestureLibraryX & \textbf{0.8602} & 0.8532 & 0.8346 & 0.8485 & 0.8589 & 0.8593\\
    UWaveGestureLibraryY & 0.8483 & 0.8438 & 0.8282 & 0.8374 & \textbf{0.8487} & 0.8455\\
    MixedShapesRegularTrain & \textbf{0.8259} & 0.8221 & 0.6856 & 0.7684 & 0.8233 & 0.8237\\
    MixedShapesSmallTrain & \textbf{0.8257} & 0.8166 & 0.7035 & 0.7512 & 0.8231 & 0.8239\\
    EOGVerticalSignal & \textbf{0.8669} & 0.8521 & 0.8283 & 0.711 & 0.8667 & 0.8639\\
    SemgHandMovementCh2 & \textbf{0.7519} & 0.7462 & 0.7006 & 0.6567 & 0.7408 & 0.7395\\
    ECG5000 & 0.7142 & 0.7047 & 0.7249 & 0.5278 & 0.7095 & \textbf{0.745}\\
    OSULeaf & \textbf{0.7583} & 0.7457 & 0.7242 & 0.7062 & 0.7549 & 0.7545\\
    Symbols & \textbf{0.8995} & 0.8934 & 0.8749 & 0.7424 & 0.8987 & 0.8986\\
    MiddlePhalanxOutlineAgeGroup & \textbf{0.7355} & 0.7281 & 0.7116 & 0.7349 & 0.7342 & 0.7347\\
    ProximalPhalanxOutlineAgeGroup & \textbf{0.8} & 0.7841 & 0.794 & 0.7722 & 0.7779 & 0.7778\\
    ProximalPhalanxTW & \textbf{0.8748} & 0.8087 & 0.807 & 0.7807 & 0.7855 & 0.7872\\
    SyntheticControl & 0.8698 & 0.8544 & \textbf{0.8797} & 0.7519 & 0.8689 & 0.8691\\
    \midrule
    Average RI & \textbf{0.8259} & 0.8135 & 0.7902 & 0.7598 & 0.8173 & 0.8193\\
    \% difference &  & -1.24\% & -3.57\% & -6.61\% & -0.86\% & -0.66\%\\
    p-value ($\alpha=0.05$) &  & 3e-4 & 3e-4 & 3e-4 & 3e-4 & 1.468e-2\\
    \bottomrule
  \end{tabular}
  \end{sc}
  \end{small}
  \end{center}
  \vskip -0.1in
\end{table*}

\begin{table*}[!tb]
  \caption{Normalized Mutual Index (NMI) ablation studies for UCR datasets. A higher NMI is better and the best results are highlighted.}
  \label{nmi-ablation-table}
  \vskip 0.1in
  \begin{center}
  \begin{small}
  \begin{sc}
  \begin{tabular}{lllllll}
    \toprule
    Dataset & LoSTer & MLP-based & RNN-based & Soft \textit{k}-means & KL & w/o CL\\
    \midrule
    CinCECGTorso & \textbf{0.2812} & 0.2414 & 0.0552 & 0.097 & 0.2569 & 0.2589\\
    NonInvasiveFetalECGThorax1 & 0.7399 & 0.685 & 0.6173 & 0.7149 & \textbf{0.7435} & 0.736\\
    NonInvasiveFetalECGThorax2 & \textbf{0.8019} & 0.74 & 0.7293 & 0.7727 & 0.7987 & 0.7981\\
    StarLightCurves & \textbf{0.6024} & 0.5851 & 0.5717 & 0.4958 & 0.6012 & \textbf{0.6024}\\
    UWaveGestureLibraryX & \textbf{0.4516} & 0.4339 & 0.3885 & 0.4001 & 0.4466 & 0.45\\
    UWaveGestureLibraryY & \textbf{0.4388} & 0.4194 & 0.3725 & 0.3973 & 0.4361 & 0.4369\\
    MixedShapesRegularTrain & \textbf{0.5147} & 0.5007 & 0.1042 & 0.3827 & 0.5081 & 0.5085\\
    MixedShapesSmallTrain & \textbf{0.5122} & 0.4941 & 0.1673 & 0.351 & 0.5071 & 0.5112\\
    EOGVerticalSignal & \textbf{0.3619} & 0.3144 & 0.1719 & 0.2012 & 0.3413 & 0.339\\
    SemgHandMovementCh2 & \textbf{0.2508} & 0.2207 & 0.1245 & 0.0298 & 0.1917 & 0.1926\\
    ECG5000 & 0.4781 & 0.4688 & 0.4593 & \textbf{0.4876} & 0.4753 & 0.4755\\
    OSULeaf & \textbf{0.2286} & 0.2185 & 0.1101 & 0.0766 & 0.2128 & 0.2115\\
    Symbols & \textbf{0.7979} & 0.7634 & 0.7359 & 0.4924 & 0.7968 & 0.7967\\
    MiddlePhalanxOutlineAgeGroup & 0.392 & 0.3933 & 0.2984 & 0.3796 & 0.3948 & \textbf{0.3954}\\
    ProximalPhalanxOutlineAgeGroup & \textbf{0.5298} & 0.5107 & 0.4986 & 0.456 & 0.4673 & 0.4666\\
    ProximalPhalanxTW & \textbf{0.6395} & 0.522 & 0.5349 & 0.484 & 0.512 & 0.5083\\
    SyntheticControl & 0.7862 & 0.7385 & 0.7315 & 0.4028 & 0.7983 & \textbf{0.8001}\\
    \midrule
    Average NMI & \textbf{0.5181} & 0.4852 & 0.3924 & 0.3895 & 0.4993 & 0.4993\\
    \% difference &  & -3.29\% & -12.57\% & -12.86\% & -1.88\% & -1.88\%\\
    p-value ($\alpha=0.05$) &  & 3e-4 & 3e-4 & 3e-4 & 3e-4 & 1.468e-2\\
    \bottomrule
  \end{tabular}
  \end{sc}
  \end{small}
  \end{center}
  \vskip -0.1in
\end{table*}

The ablation studies serve to verify the contribution of the proposed dense autoencoder and optimization strategy: 1) LoSTer with the MLP backbone of IDEC, DNFCS, CKM, (2) LoSTer with the RNN backbone of DTCR and DTCC, (3) LoSTer optimized through soft \textit{k}-means, (4) LoSTer optimized through the KL divergence, (5) LoSTer without contrastive learning.

Tables \ref{ri-ablation-table} and \ref{nmi-ablation-table} show LoSTer scoring superior RI and NMI on average across the various datasets to most of its ablations. The highest average increments are +3.57\% and +12.57\% of RI and NMI against the RNN-based version, while +6.61\% and +12.86\% of RI and NMI against the same neural network optimized through soft \textit{k}-means. Therefore, we observe a large margin of improvement due to replacing the RNN architecture and to the \textit{k}-means differentiable optimization. The contribution of the LoSTer autoencoder compared to the MLP baseline by IDEC, DNFCS, and CKM on the one hand and of dual contrastive learning, on the other hand, are more limited but still beneficial. Importantly, LoSTer performs better under the \textit{k}-means differentiable cost function as it outperforms both the soft \textit{k}-means and KL-divergence variants. The significance of the difference of each method against LoSTer is confirmed by the Wilcoxon signed rank test at the 0.05 p-value level.

\subsection{Large-Scale Time Series Clustering}
\begin{table*}[!htb]
  \caption{Rand Index (RI) and Normalized Mutual Index (NMI) for real-world retail datasets. \\Higher is better and the best results are highlighted.}
  \label{large-scale-table}
  \vskip 0.1in
  \begin{center}
  \begin{small}
  \begin{sc}
  \begin{tabular}{lllllllll}
    \toprule
    Dataset & Metric & LoSTer & IDEC & DNFCS & CKM & DTC & DTCR & DTCC\\
    \midrule
    \multirow{ 2}{*}{M5} & RI & \textbf{0.8985} & \textbf{0.8985} & 0.8973 & 0.893 & 0.8914 & 0.8973 & 0.898\\
    & NMI & \textbf{0.1836} & 0.1705 & 0.1689 & 0.1678 & 0.111 & 0.1288 & 0.1368\\
    \midrule
    \multirow{ 2}{*}{Store Sales} & RI & \textbf{0.9523} & 0.9281 & 0.9247 & 0.915 & 0.9199 & 0.9116 & 0.9347\\
    & NMI & \textbf{0.4844} & 0.4298 & 0.4233 & 0.4216 & 0.2663 & 0.2354 & 0.2576\\
    \bottomrule
  \end{tabular}
  \end{sc}
  \end{small}
  \end{center}
  \vskip -0.1in
\end{table*}

In addition to common long-term benchmarks, we assess LoSTer on two challenging large-scale real-world retail forecasting datasets.

1) The M5 competition dataset contains 30490 time series with 1913 time steps recording the daily sales of items of different categories at Walmart in various stores, departments, and states over 6 years in the USA. However, M5 is a sparse dataset: since the many zero-sales days may impact clustering more than other sales characteristics, only 1150 time series are considered and we aim to group them into 29 clusters according to category and store.

2) The Store Sales competition dataset is used for predicting daily sales of thousands of products sold by different stores in the main cities of Ecuador over 5 years. Here, the scope is clustering 1729 time series of 1684 steps into 33 families of products.

RI and NMI scores are reported in Table \ref{large-scale-table}. Even though all models are comparable on M5 in terms of RI, LoSTer achieves a 0.1836 NMI score, which is higher than the 0.1705 of the runner-up IDEC. The margin is greater for the Store Sales dataset: the RI score by LoSTer is 0.9523 and higher than DTCC at 0.9347, while the NMI score is 0.4844 against 0.4298 by IDEC. Other models are less accurate.

\subsection{Training and Inference Speed}
\begin{figure}[!b]
  \centering
  \includegraphics[width=0.97\columnwidth,keepaspectratio]{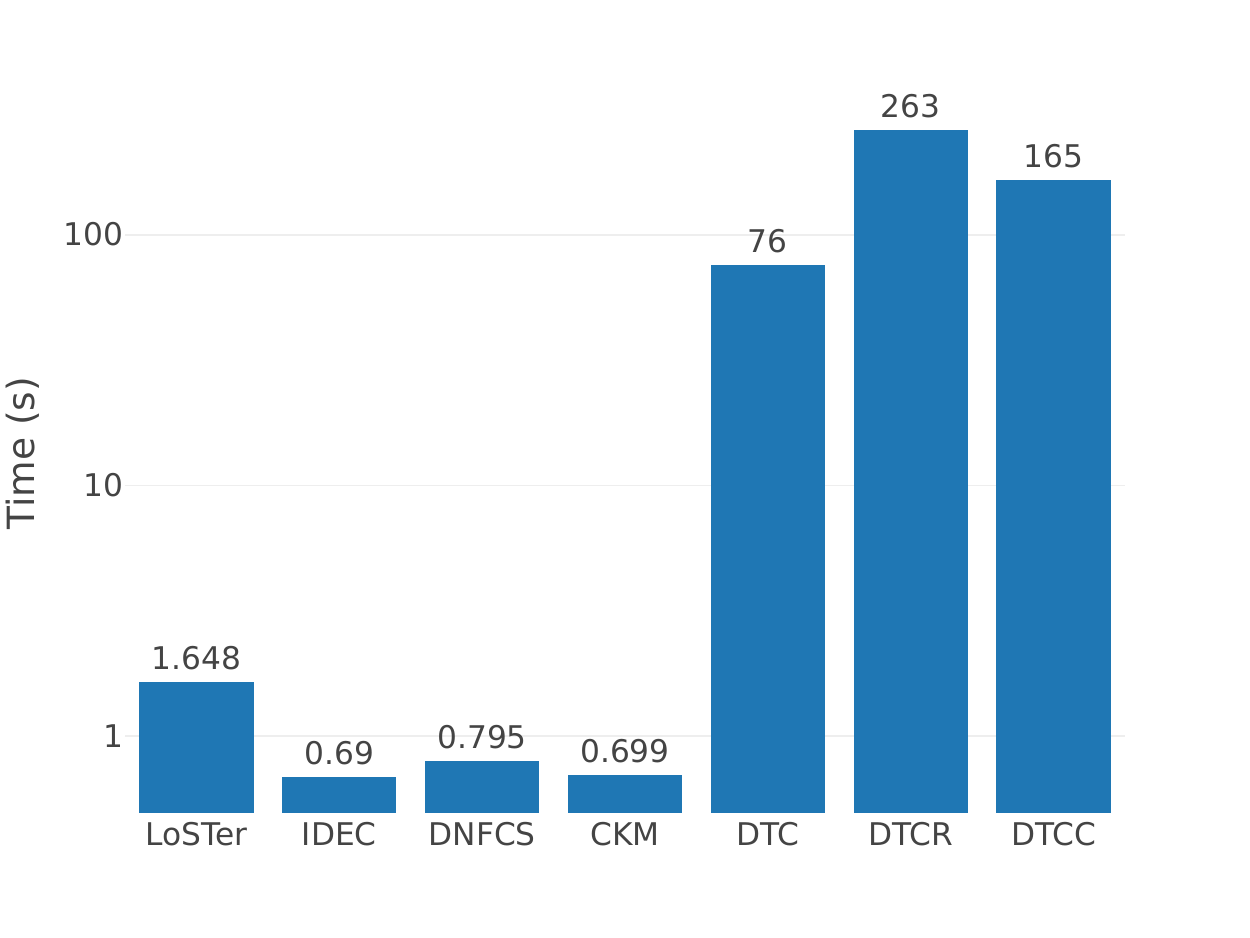}
  \caption{The bar chart reports the actual training times [s] by methods for one epoch referred to the Store Sales dataset. The y-axis is plotted in log-scale. The fastest IDEC and slowest DTCR are distanced by 3 orders of magnitude.}
  \label{fig:training_efficiency}
\end{figure}

\begin{table}[!htb]
  \caption{Training and inference time for the StarLightCurves dataset.}
  \label{time-table-transformers}
  \vskip 0.1in
  \begin{center}
  \begin{small}
  \begin{sc}
  \begin{tabular}{llll}
    \toprule
    Model & Train & Inference\\
    \midrule
    LoSTer & 2.85 s & 1.81 s\\
    Preformer & 1531 s & 270 s\\
    Pathformer & 550 s & 60 s\\
    iTransformer & 3.66 s & 2.03 s\\
    \bottomrule
  \end{tabular}
  \end{sc}
  \end{small}
  \end{center}
  \vskip -0.1in
\end{table}

MLP-based models enjoy faster training time than RNNs whose computations must be performed sequentially because the next output depends on the hidden state from the previous iteration. The single-layer RNN in the decoding phase of DTCR and DTCC limits severely the computing speed of the GPU, especially on long sequences. We demonstrate these effects by reporting in Fig. \ref{fig:training_efficiency} the average training times for one epoch in seconds on the Store Sales dataset with batch size 32. LoSTer completes one epoch in less than 2 seconds, which is lower by far than DTC (76 s), DTCR (263 s), DTCC (165 s) and still close to MLP competitors: IDEC (0.69 s), DNFCS (0.795 s), CKM (0.699 s). The simple yet effective dense architecture of LoSTer exhibits faster training than recurrent models. All running times are measured on a machine powered by a single NVIDIA A4000 GPU and 24 cores Intel(R) Xeon(R) CPU @ 2.40GHz.

Table \ref{time-table-transformers} reports the same time comparison for Transformer-based models in completing one training and one inference data pass of the StarLightCurves massive dataset. While the training and inference by LoSTer and iTransformer take around 3 seconds, Preformer and Pathformer require hundreds of seconds or more because of learning and then attending the many temporal tokens of input time series. Eventually, the RNNs and Transformers' computational complexity leads to longer training/inference times - in addition to poor clustering accuracy - making their real-world deployment infeasible for LSTC purposes.

\section{Conclusion}
We have proposed LoSTer, a novel deep learning method designed for long-sequence time series clustering (LSTC) solving the non-differentiability of the canonical \textit{k}-means objective through the Gumbel-softmax reparameterization trick, and boasting more accurate and faster training than recurrent and Transformer-based models through a dense residual contrastive architecture. Specifically, LoSTer overcomes the issues of sub-optimal clustering, error accumulation and slow training caused by surrogate losses and autoregressive reconstruction. We have also provided theoretical discussions based on the known findings from the long-term time series forecasting research about the inadequacy of the attention in Transformers to effectively capture temporal dependencies over long horizons while also being hindered by high computation and space requirements. These issues are exacerbated in LSTC because time series are clustered at the batch level, hence the clustering performance can deteriorate by falling back to a small batch size.
Extensive experiments on benchmark and real-world datasets validate the optimization strategy and model's dense architecture which enable LoSTer to outperform the existing RNNs and the state-of-the-art Transformers by a significant margin in terms of accuracy and training/inference time, thus eventually delivering a simple, fast to train and scalable method for LSTC.

\bibliographystyle{IEEEtran}
\bibliography{biblio}

\end{document}